\pdfoutput=1

\documentclass[11pt]{article}

\usepackage{emnlp2021}

\usepackage{times}
\usepackage{latexsym}

\usepackage[T1]{fontenc}

\usepackage[utf8]{inputenc}

\usepackage{microtype}

\usepackage{amsmath,amssymb}
\usepackage[flushleft]{threeparttable}
\usepackage{array,booktabs,makecell}
\usepackage[font=small]{caption}
\usepackage{todonotes}
\usepackage{tabularx}
\usepackage{booktabs}

\title{A Survey of Online Hate Speech through the Causal Lens}

\author{Antigoni-Maria Founta \and Lucia Specia\\
  Department of Computing \\
  Imperial College London \\
  United Kingdom \\
  \texttt{\{a.founta20,l.specia\}@imperial.ac.uk}}

\begin{document}
\maketitle
\begin{abstract}

The societal issue of digital hostility has previously attracted a lot of attention. The topic counts an ample body of literature, yet remains prominent and challenging as ever due to its subjective nature. We posit that a better understanding of this problem will require the use of causal inference frameworks. This survey summarises the relevant research that revolves around estimations of causal effects related to online hate speech. Initially, we provide an argumentation as to why re-establishing the exploration of hate speech in causal terms is of the essence. Following that, we give an overview of the leading studies classified with respect to the direction of their outcomes, as well as an outline of all related research, and a summary of open research problems that can influence future work on the topic.
\end{abstract}

\section{Introduction}

User-generated content has flourished with the evolution of social media platforms, and with it there has been an emergence of several social phenomena. Some of them can have a positive impact on mental health and eventually prove beneficial for societies; for example, it has been shown that the use of social networks from individuals of advanced age can lead to social benefits as well as have promising cognitive effects \cite{quinn2018cognitive}. There is, however, a plethora of societal issues on the digital world which are proliferating in these platforms; such are several types of online misbehaviours.

Despite the clear interest of the research community to prevent, detect and filter harmful content \cite{schmidt2017survey, fortuna2018survey, macavaney2019hate, mishra2019tackling, tontodimamma2021thirty, vidgen2019challenges}, the problem is very complex and still far from addressed. For instance, its multifactorial character poses already a large obstacle with manifold parameters to consider: diversity of platforms, abundance of  languages, various facets it can manifest on, and different forms of language, to name a few. Even at a higher level, there are some deeply philosophical issues to contemplate, such as the thin line between awry, unwelcome content and freedom of speech. Nonetheless, the impact of online communications on society is considered more beneficial than harmful, hence it is important to advance the current scientific work to address its limitations.

The matter of unhealthy discourse is widely context-dependent \cite{saleem2017web} and subjective \cite{salminen2018online, aroyo2019crowdsourcing}. A racial slur, for instance, will be interpreted differently from individuals with separate ethnic or racial backgrounds. Conversely, however, most existing approaches operate on the individual post/comment level or aggregate isolated pieces of information -- usually of textual nature -- and seek simplistic, ostensibly universal, decisions. Even though this was a necessary course of action at the beginning, such nuanced task requires higher sophistication. To combat the issue to the core, it is of utmost importance to deeply understand the factors that drive it and determine them in a definite way. 

An increasingly popular scientific approach to ensure certainty is via the application of causal methods. Causality suggests an area of study that is ubiquitous in many parts of modern science \cite{russo2011epistemic, young2016unifying, landes2018epistemology, hicks1980causality}, among which 
social sciences \cite{marini1988causality, reiss2009causation}. Especially with the development of methodologies to extract causation from observational, instead of experimental, studies, it is now possible to utilise the profusion of web content for the extraction of causal knowledge \cite{de2017language, saha2019social, doan2019extracting}.

Driven from the necessity of contemplating hate speech from a causal prism, there exists a relevant, yet still narrow, body of literature. In this survey, we attempt to summarise the existing research which analyse phenomena of digital misbehaviours from a causal perspective, quantifying the influence on fundamental sociological outcomes related to harmful language online. In addition, we highlight the important research gaps and make suggestions for future directions. The goal of the present paper is to call attention to the importance of causation in the context of the discussed issue. In addition, it is intended to act as a point of reference for aspiring practitioners in the topic.

The rest of the paper revolves around the compilation of relevant publications, first classified based on their focus of sociological impact (\S\ref{sec:sociological}) and, subsequently, arranged according to their approach towards salient methodological settings. The latter section (\S\ref{sec:technical}) is organised into several prominent hurdles, which include information regarding how they are treated from the reviewed work and points of discussion about the research gaps.
\section{Sociological Causal Impact}
\label{sec:sociological}

Digital profanity is, first and foremost, a sociological phenomenon that affects many aspects of the virtual world. There are some pronounced directions which require thorough examination, such as the consequences of such actions on the affected communities as well as their targets, or the impact of banning policies aiming at fighting against it. Moreover, there are several other underlying effects that could get in the spotlight, including the ramifications of interventions or the reasons that drive toxic actors. As a result, it is of paramount importance to examine hate speech in a holistic way and concretely quantify the drivers of these outcomes. However, it is impossible to achieve this task without considering causality, because non-causal inferences can never be conclusive. Surprisingly, despite the broad interest of the research community in this topic, very little work has been done on attempting causal links, even on the most prominent tasks related to online hate speech (OHS). 

In the present survey, we classify fundamental sociological outcomes related to OHS and outline the most distinguished body of literature. The classification results into three major pillars, with respect to the following:
\begin{itemize}
    \item \textbf{Digital misbehaviours versus the physical world}: we summarise studies concerning the propagation of online hate speech to real life \cite{muller2018fanning, muller2020hashtag} as well as the influence of offline events to the dissemination of the issue online \cite{olteanu2018effect, thomas2021behavior}\footnote{This direction has also been studied by \cite{scharwachter2020does}, however they merely calculate correlations and do not control for potential confounders, hence are not included in the present survey.}.
    \item \textbf{Harmful content versus the individuals}: we outline research concerning the impact of toxic behaviours on the targets or passive readers \cite{saha2019prevalence} as well as the by-products of web characteristics (such as anonymity) on hate speech producers \cite{von2020misogynistic}.
    \item \textbf{Effect of interventions}: finally, we review works that revolve around quantifying the effect of combating strategies which various platforms adopt. Existing research has focused on limiting policies, censoring and counter-speaking \cite{alvarez2018normative}, social sanctioning \cite{munger2017tweetment}, quarantining \cite{chandrasekharan2020quarantined} and banning hateful communities \cite{chandrasekharan2017you, thomas2021behavior}.
\end{itemize}
In all the of the cases, we can consider the role of hate speech both from the position of a phenomenon which is the result of (potentially) multiple causes, but also from the position of the causal root and study its effects. Therefore, in the following sections we review both directions.

\subsection{Digital and Physical World}

The first and most apparent pillar concerns the interrelation of online hate speech with the physical world, in a range of simple dissemination to absolute influence. Towards this direction, \citet{muller2018fanning, muller2020hashtag} have conducted research on the causal effect of social media on the propagation of hate sentiments offline, whether anti-refugee \cite{muller2018fanning} or more specifically anti-Muslim \cite{muller2020hashtag}. In \cite{muller2018fanning}, the authors provide evidence that there is significant association between negative content against refugees existing on Facebook and offline hate crimes in Germany on a municipal level, while controlling for multiple potentially confounding factors such as German municipalities' characteristics and overall social media usage. To reach to this conclusion they combine a variety of data sources; online anti-refugee sentiment is represented by content from a widespread Facebook page of a German right-wing party, which hosts plenty of far-right content; controlling for the network's popularity in Germany, the authors measure Facebook outages and internet disruptions; finally, to further measure user activity on the network and create controls based on a neutral subject, they explore another broadly popular page of a famous commercial product. 
The causal framework they implement is a fixed-effects regression model (inspired by \citet{bartik1991boon}), which considers the aforementioned panel data combined with a range of controls. They discover that the effect is stronger in areas with higher Facebook usage and demonstrate a robustly strong connection between the activity of the right-wing group and severe hate crimes.

Similarly, in \cite{muller2020hashtag}, they study the causal impact of Islamophobic social media content on registered crimes and overall negative sentiment against Muslims, and whether former US president Donald Trump's Twitter campaign has contributed to the propagation of Islamophobia. To ensure validity and robustness of their findings, they fuse a number of different data sources and employ a difference-in-differences approach. The data originate mainly from Twitter for the social media information, a survey by FBI to discover hate crimes, data from mass media, demographic information about US counties etc. Their findings provide evidence to associate a 38\% larger increase in hate crimes, between 2010 and 2017, with higher exposure to social media. Moreover, consistently with previous research, they also provide evidence which shows a connection amid the start of Trump's presidential campaign with an increase of anti-Muslim sentiments in USA. Both projects illustrate there is strong evidence linking OHS with offline occurrences of hate-related crimes, with the former having a causal effect on the latter. \footnote{It goes without saying that results from papers that employ causality need to be interpreted with great care. As \citet{muller2018fanning} emphasise, for instance, their findings do not indicate that social media can cause crimes against minorities, but rather "\textit{social media can act as a propagating mechanism [...]}" so that "\textit{shifts in exposure to anti-refugee sentiment on social media can increase the number of anti-refugee attacks}".}

Looking at the opposite direction, online hate speech is frequently affected by events taking place in the offline world. For example, following the September 11\textsuperscript{th}, 2001 attack to the twin towers of New York City, there seems to be an increase in Islamic terrorist attacks as much as an increase in Islamophobia. It is very well expected that this will affect the online world and the way Muslims are perceived, which seems to actually be the case according to \citet{olteanu2018effect}. In \cite{olteanu2018effect} they analyse the influence of offline events on online hate speech. More specifically, they study the impact of several attacks related to the Islamic State -- both Islamic terrorist attacks and Islamophobic ones -- in terms of online hate speech and counter-hate speech. Their findings indicate that terrorist attacks show an increase in OHS, especially towards Muslims and Arabs. 

To calculate the causal effect, they construct time series from Reddit and Twitter -- representing the number of posts and unique users involved with the event -- and then synthesize counterfactual time series for the same period of time, such as they would be produced had the attacks not happened. The counterfactual data are created by composing timelines of the same event, while adjusting for a temporal shift prior to the event, so that the time series will reflect a similar period of time but in different time windows. The produced timelines, put together with other external data sources, are then fit into a state space model \cite{brodersen2015inferring} using maximum likelihood estimation, to predict the synthesized control time series. Comparing the treatment and control series, they calculate relative effects for a number of manually curated terms, which are also annotated across four hate speech dimensions (stance, targets, severity, and framing). 

\subsection{Actors and Targets}

Beyond broadly looking at the overall impact of OHS, the next step would be to concentrate on the individuals and speculate how OHS affects and is affected by the participating members, whether these are at the producing or receiving end of such content. For example, in the aforementioned works, \citet{olteanu2018effect} and \citet{muller2020hashtag} have made some general remarks regarding OHS, but in order to effectively focus on the task they narrow the type of hate speech to be racism and, more specifically, Islamophobism. Emphasis, however, is given on understanding the dynamics of diffusion and not on studying the impact on individuals who support Islam. 

On the these grounds, \citet{saha2019prevalence} study the effect of hate prevalence on the stress levels of US college students, within Reddit college communities. To measure the levels of toxicity, they employ hate lexicons, with the help of which they compute the College Hate Index (CHX), as fractions of hateful keywords in each community compared to other subreddits banned for violating the hate-limiting policies of Reddit. Similarly, they quantify the exposure of users to hateful content based on the threads they have participated and to account for their stress levels they use a binary classifier based on existing models. They examine numerous observable confounders -- such as the subreddit and user activity -- and apply propensity score matching to calculate the causal effect whilst controlling for the covariates. Their results demonstrate an increase in stress expression caused by exposure to hateful speech.

Additionally, inherent characteristics of online environments, such as anonymity, ease of access, and size of audience \cite{brown2018so} are highly likely to affect the behaviour of online social networks' users and sometimes make it easier for them to misbehave. \citet{von2020misogynistic} discuss the outcomes of the by-products of online world characteristics -- in this case anonymity -- on hate speech actors. In particular, they compare the degree of hatefulness before and after the identities of a large set of users from Flashback, an anonymous Swedish discussion platform similar to Reddit, have been publicly exposed. Their hypothesis suggests that once running the risk of exposure, users decrease the volume of hateful content they post. To detect hate, they implement a machine learning model and make predictions on the data, which will afterwards be used with a difference-in-differences approach to make causal claims. According to their estimates, the reduction of anonymity, as in risk of exposure, leads to a decline in general hate and overall activity, and even more on xenophobic content. Surprisingly, levels of misogyny increase. These empirical findings mostly support the author's original hypothesis. 

\subsection{Interventions}

Last but not least, perhaps the most extensively studied of all three pillars is the effect of interventions, possibly due to its relatively close connection to causality. Interventions here refer to actions that are taken towards the elimination of the phenomenon; such strategies include quarantining \cite{chandrasekharan2020quarantined}, banning \cite{chandrasekharan2017you, thomas2021behavior}, censoring \cite{alvarez2018normative}, sanctioning \cite{munger2017tweetment}, and counter speaking \cite{alvarez2018normative}. Reverberations of each policy vary, depending on the platform and the methodology followed.

\citet{chandrasekharan2017you} first studied the effect of banning an entire hateful community using causal inference methods. More specifically, they investigate through a quasi-experiment how banning targeted communities influenced hate speech levels on Reddit. Initially, they investigate the activity of the participants, post-banning the examined subreddits, with respect to activity level and hateful content volume. To control for potential confounders related to user characteristics -- such as the activity, popularity, or age of user accounts -- they employ Mahalanobis Distance Matching \cite{rubin2006affinely} between treatment and control users, which is then further enhanced with a difference-in-differences analysis \cite{abadie2005semiparametric} of the two groups over time. Subsequently, they inspect the level of hate propagation to other \textit{invaded} (as they call them) subreddits. In this case, matching needs to be applied on a subreddit level, rather than user level, so instead they employ the Interrupted Time Series approach \cite{bernal2017interrupted}. Their results show that the ban worked \textit{for Reddit}, meaning that having targeted a particular area of the platform, they have successfully eliminated hateful content without conveying the problem elsewhere within Reddit. 

In addition to this work, the authors also studied the causal effect of \textit{quarantining} Reddit communities, in a similar experimental framework \cite{chandrasekharan2020quarantined}. Quarantining is a form of intervention that Reddit applies, where communities are indicated as potentially problematic and users have to deliberately choose to enter them, after being warned about toxicity levels within. This approach is less stern than banning, however according to the findings of this study, it is still effective to restrict the influx of users in hostile communities, while preserving the freedom of speech. In this case, the causal inference strategy used is the Interrupted Time Series regression \cite{bernal2017interrupted}, which models interruptions caused by the treatment variable.
\begin{table*}[ht]
\small
\begin{tabularx}{\textwidth}{XXXXXX}
\toprule
\textbf{Paper} & \textbf{Domain} & \textbf{Treatment} & \textbf{Outcome} & \textbf{Controls} & \textbf{Causal Method}  \\ 
\toprule
    \citet{chandrasekharan2017you} & Reddit & Banning subreddits & Former members' activity, language \& migration trends & Subreddits that could potentially have been banned & MDM, DiD \& ITS Regression Analysis \\ \hline
    \citet{munger2017tweetment} & Twitter & Social sanctioning, varying by influence \& identity & Race-based harassment & Gender \& race of harassers, and anonymity & Randomized Control Experiment \\ \hline
    \citet{olteanu2018effect}  & Twitter \& Reddit & Islamophobic \&  Islamist terrorist attacks in Western countries & Hateful content & Synthetic counterfactual time series & Comparison of observed vs counterfactual time series \\ \hline
    \citet{alvarez2018normative}  & Custom forum & Censoring \& counter-speaking & Hate speech score & Number of comments & Experimental approach \\ \hline
    \citet{muller2018fanning}  & Facebook & Anti-refugee Facebook group & Hate crimes against refugees in Germany & Characteristics of German municipalities & Fixed effects panel regressions \\ \hline
    \citet{saha2019prevalence}  & Reddit & Hateful speech in college subreddits & Online stress levels & Subreddit \& User activity & Propensity score matching \& DiD Regression Analysis \\ \hline
    \citet{muller2020hashtag}  & Twitter & High Twitter usage & Anti-Muslim hate crimes & Characteristics of US counties & DiD approach \\ \hline
    \citet{von2020misogynistic}  & Flashback & Web anonymity & Hateful posts & Risk of web exposure & DiD approach \\ \hline
    \citet{chandrasekharan2020quarantined}  & Reddit & Quarantining subreddits & Member participation \& language, and new member influx & Subreddits that could potentially have been quarantined & ITS \& Bootstrapping tests \\ \hline
    \citet{thomas2021behavior}  & Reddit & External events \& regulatory actions (ie. bans) & Member activity \& attrition & User participation & MDM \& DiD analysis \& Multivariate Bayesian changepoint analysis \\ \hline
    \citet{ananthakrishnan2021drivers} & YouTube & Presence of hate speech & Propagation (or "virality") of hate speech & Popularity, quality and posting time of videos & Instrumental Variables \\ \hline
\toprule
\end{tabularx}
\caption{\label{citation-guide}
Following the format of \cite{keith2020text}, this table summarises research papers related to causal effects of web hostility. The abbreviations used stand for the following; Difference-in-Differences (DiD) \cite{abadie2005semiparametric}; Interrupted Time Series (ITS) \cite{bernal2017interrupted}; Mahalanobis Distance Matching (MDM) \cite{rubin2006affinely}.
}
\label{tab:summary}
\end{table*}
\section{Difficulties and Causal Methods}
\label{sec:technical}

Experimentation for the sole purpose of causality is fairly costly and, in this case, can be largely unethical, considering that it would mean deliberate exposure of people to toxic material. As a result, it is crucial to exploit the available methodologies of observational causality, in combination with its intersection with natural language processing. There are, however, several open research problems which lie at the heart of textual causality and thus are inherited by this task. A prominent hurdle, for instance, exists in the conception of the causal diagram, contemplating all the possible covariates and determining their roles. Such graphs are domain-specific and can greatly deviate, due to differences introduced by the platforms or  niche types of speech. Furthermore, there are challenges such as the discovery of effective linguistic representations for the various parts where text can act as a surrogate. 

In this section we discuss some prevailing difficulties, summarise the route of relevant literature, and finally, provide some suggested directions based on previous studies.

\subsection{Confounding Bias}

To establish true causation among a target variable X and an outcome Y, there should not be any indirect connections influencing the effect of X on Y. In reality it is rarely the case that this will hold organically true, hence being aware of any confounding factors is fundamental in order to intervene accordingly and eliminate their impact. Many confounders can be detected after careful exploration of the domain and quantified through unadjusted versus adjusted estimates. These are called \textit{observable} confounders and are largely discovered based on domain expertise or previous research. It is well possible, however, that not all such factors can be anticipated; there are cases where they might not be discoverable or their computations might not be feasible.

One very relevant work on treating \textit{latent} confounders is the one of \citet{cheng2019robust}, who attempt to create a somewhat causally-aware cyberbullying detection model for observed data, making it robust on confounding bias by controlling for latent (or plausible, as they call them) confounders. To achieve that, they look for pairs of variables that demonstrate Simpson's paradox and then employ a clustering algorithm to group the data based on the discovered p-confounders. They afterwards perform classification within the clusters. Their proposed framework of detecting latent confounding factors is potentially generalisable outside the scope of this topic. Nevertheless, to the best of our knowledge, this is the only research project considering cyberbullying detection from a causal prism. 

\paragraph{Suggestions}

While controlling for confounders, and in case they are categorical or numerical variables, it is possible to follow well established techniques such as matching \cite{de2011matching}. In simple terms, matching is a method to create pairs of samples from the same category and of different outcomes, to approximate randomised conditions and obliterate confounding bias so that any effect that remains must be realistic. For example, \citet{zhang2018conversations} perform their task while controlling for topical confounding, by creating pairs of "good" and "bad" conversations from the same Wikipedia page. This way, they ensure that any differences caused by the nature of the Wikipage topic are removed. However, it is not always possible to know the confounding factors. In that case, it is more useful to consider ways of eradicating confounding in a holistic way, such as the clustering approach of \cite{cheng2019robust}.

Additionally, it is highly likely for features of this language-dominated topic to be textual. In that case, treating the covariates can be fairly challenging. Drawing inspiration from the textual causality literature \cite{keith2020text}, there are a number of approaches both for observable and latent textual confounders. For the former, it is possible to follow the strategies that will be described in Section \ref{linguistic_representations} and extract features that best represent each specific factor. The case of latent confounders, though, is more elaborate. One suggestion proposed by \citet{landeiro2019discovering} is the use of adversarial learning as a combating method to confounding shift, although it has yet to be examined within the context of OHS.

\subsection{Linguistic Representations}
\label{linguistic_representations}

Most of the  body of literature described in Section \ref{sec:sociological} base their research on numerical features, regression, and time series analysis and do not look at the actual content of the social media platforms or question the validity of hatefulness. Contrary to traditional approaches to hate speech detection, which mostly work with text, these causal analyses largely overlook language. One possible explanation for this could be the inherent difficulty of constructing representative vectors that will capture the verbal essence. Language as a means of communication is intrinsically high-dimensional and reducing it in lower dimensions to allow for processing requires strenuous effort.

\citet{saha2019prevalence} and \citet{cheng2019robust} are the only previous papers, to the best of our knowledge, to exploit the texts and attempt to represent them with some feature vectors, to produce causal representations of OHS: frequency of hate keywords in the first case, and Linguistic Inquiry and Word Count (LIWC --\citet{tausczik2010psychological}) in the second. There is great need for experimentation and improvement to attempt more holistic approaches to the problem.

\paragraph{Suggestions}
Since this task is significantly understudied in the context of online hate speech, there are a lot of possible directions to follow. For example, it might prove meaningful to extract high-level representations like sentiment features or toxicity scores. The latter is supported by  \citet{von2020misogynistic}, who claim that "hate begets hate", meaning that hateful content triggers replies of the same style. Furthermore, one could use the framework implemented by  \citet{pryzant2018deconfounded}, which automatically induces representative lexicons for social science tasks, while controlling for potential confounding factors. Having previously established some factors as such, it is possible to employ this framework and observe the performance of the lexicon. Lastly, it could prove fruitful to explore pre-trained language representation models that are focused on capturing profanity (e.g. HateBERT \cite{caselli2020hatebert}) or causality (e.g. CausalBERT \cite{khetan2020causal, veitch2020adapting}), or even attempt to fine-tune a hate- and causal-specific version of the original BERT model \cite{devlin2018bert}, for a combination of the two. The latter would probably be very demanding with respect to the need for training data, albeit promising.

\subsection{Causal Transportability}
Finally, a big benefit of true causal knowledge is the ability to transfer this knowledge across domains and achieve comparably optimal performance \cite{pearl2011transportability}. Such an achievement can then work as an evaluation method, showing that the exhibited causal attributions are, in fact, correct and sufficient. For example,  consider racist material, which is possibly similar across social networks. A detection model that takes into account  any previously attributed causal factors, and controls for all discovered confounders, should be able to train with data from one network and exhibit satisfactory performance on a different network. \cite{cheng2019robust} is the only existing study that is attempting to address causal transferability in the context of this topic, by experimenting on one domain and testing on a different, switching between Twitter and Formspring. Despite the challenging task, it would be of great interest to apply this evaluation method across alternate platforms, various labels, or even multiple languages.
\section{Conclusions}

The task of hate speech mitigation can never be conclusively accomplished unless the research community seriously speculates about its roots. This would mean a methodological experimentation with the phenomenon, from a causal prism. Simply put, it would mean taking a step back from its detection and focusing on scientifically breaking it into its causes. If the causes are known, then any further steps can make meaningful change. Practitioners should then be able to take crucial actions towards either its efficient and interpretable detection or, even further, its {\em a priori} prevention. By understanding the most prominent and fundamental factors leading to the phenomenon, it is possible to conclusively build on its holistic solution.

The present survey is intended to initiate discussions towards this direction. By introducing some of the most relevant studies while simultaneously highlighting several points of interest, we aspire for this paper to work as a point of reference, as much as an inspiration and an open call for further research. To make any definite causal claims it is necessary to have a well defined system, where all influential factors are understood or known. All the existing research, including this survey, can only serve as a prelude to future discussions regarding this ubiquitous issue, which will consequently lead to more sophisticated and generalisable models.

\bibliographystyle{acl_natbib}
\bibliography{main}

\end{document}